\documentclass[11pt]{article}

\usepackage{ACL2023}
\usepackage{times}
\usepackage{latexsym}
\usepackage{lipsum}
\usepackage{amsmath}
\usepackage{amssymb}
\usepackage{amsfonts}
\usepackage{algorithm}
\usepackage{algorithmic}
\usepackage{graphicx}
\usepackage{multirow}

\usepackage[T1]{fontenc}

\usepackage[utf8]{inputenc}

\usepackage{microtype}

\usepackage{inconsolata}

\title{Replicating Complex Dialogue Policy of Humans via Offline Imitation Learning with Supervised Regularization}
\author{
    Zhoujian Sun$^1$, Chenyang Zhao$^1$, Zhengxing Huang$^2$, Nai Ding$^{1,2}$ \\
    $^1$Zhejiang Lab, $^2$Zhejiang University \\
    \texttt{\{sunzj, c.zhao\}@zhejianglab.com} \\
    \texttt{\{zhengxinghuang, nai\_ding\}@zju.edu.cn}
}

\begin{document}
\maketitle
\begin{abstract}
Policy learning (PL) is a module of a task-oriented dialogue system that trains an agent to make actions in each dialogue turn. Imitating human action is a fundamental problem of PL. However, both supervised learning (SL) and reinforcement learning (RL) frameworks cannot imitate humans well. Training RL models require online interactions with user simulators, while simulating complex human policy is hard. Performances of SL-based models are restricted because of the covariate shift problem. Specifically, a dialogue is a sequential decision-making process where slight differences in current utterances and actions will cause significant differences in subsequent utterances. Therefore, the generalize ability of SL models is restricted because  statistical characteristics of training and testing dialogue data gradually become different. This study proposed an offline imitation learning model that learns policy from real dialogue datasets and does not require user simulators. It also utilizes state transition information, which alleviates the influence of the covariate shift problem. We introduced a regularization trick to make our model can be effectively optimized. We investigated the performance of our model on four independent public dialogue datasets. The experimental result showed that our model performed better in the action prediction task.
\end{abstract}

\section{Introduction}
Policy learning (PL) is a module of a task-oriented dialogue system that trains an agent to make actions in each dialogue turn, where each action represents the topic of the following system utterance \cite{ni-2022-recent}. Recent studies usually adopted the reinforcement learning (RL) framework to train a dialogue agent. These studies generally regarded a dialogue as a multi-round question-answer process, and the agent is responsible for asking questions to users (e.g., EAR, UniCorn \cite{lei-2020-estimation,deng-2021-unified}) or response prompts from users (e.g., instructGPT, chatGPT\footnote{instructGPT and chatGPT are end-to-end that combine all components of a dialogue system in a single model} \cite{ouyang-2022-training,openai-2022-chatgpt}) appropriately to obtain maximum rewards. These studies require simulators to train agents \cite{sutton-2018-reinforcement}.

\begin{figure}[tb]
  \centering
  \includegraphics[width=7.5cm,height=7.03cm]{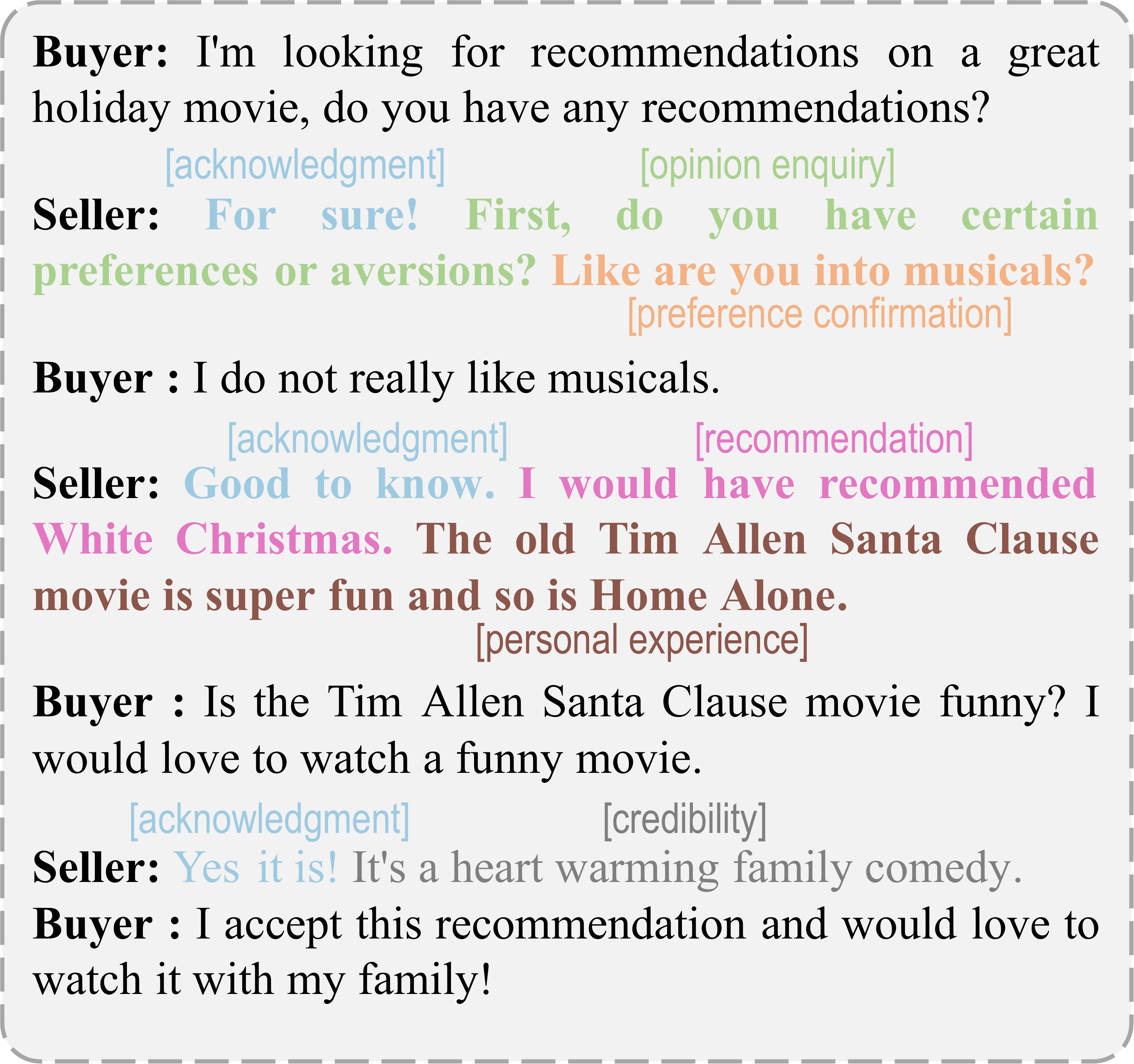}
  \caption{A conversation snippet between a buyer without an explicit goal and a seller with sociable actions.}
  \label{fig:dataSample}
\end{figure}

However, real dialogues are not limited to multi-round question-answer processes. For example, when a seller needs to recommend a movie to a buyer, the buyer may not know what they want to buy. The seller needs to lead the conversation proactively and recommend items even if they are unfamiliar with the buyer's preferences. The seller must also adopt social skills such as acknowledging and sharing personal experiences to persuade the buyer to accept the recommended item (see Fig. \ref{fig:dataSample}, \cite{hayati-2020-inspired}). Training RL-based agents to learn such complex policies requires interacting with a simulated buyer who can behave like real humans, which is beyond current technology's ability. Therefore, we argue that the RL-based agent cannot behave like a human in many complex dialogue scenarios \cite{liu-2020-towards}.

This study aims to explore how to train an agent that can imitate the complex policy of humans. A straightforward approach would be to use the supervised learning (SL) method \cite{wang-2020-multi,wang-2021-modelling}, which theoretically can learn arbitrary complex dialogue policy with annotated data. However, the practical performance of SL-based models is restricted because of the \textit{covariate shift} problem \cite{brantley-2019-disagreement,merdivan-2017-reconstruct}. Specifically, a dialogue is a sequential decision-making process where slight differences in current utterances and actions will cause significant differences in subsequent utterances. Under this circumstance, training and testing dialogue data gradually violate the independent and identically distributed (i.i.d.) assumption with the development of dialogues, leading the chance of SL-based models making mistakes to grow quadratically with dialogue turns increasing \cite{ross-2009-efficient}.

The main contribution of this study is that we proposed an offline imitation learning (OIL) based model named Supervise regularized Distributional correction estimation (SD) that alleviates the effect of covariate shift problem and learns dialogue policy better. Specifically, The SD model formulates a dialogue as a Markov Decision Process (MDP) and utilizes state-action-next-state tuples to train the agent. \citet{ross-2011-reduction} proved that utilizing the sequential state transition information can decrease the mistake growth speed from quadratic to linear, bringing performance improvement potential. The SD model learns policy from real annotated dialogue datasets, making it can learn complex human policy in an offline manner. Furthermore, we introduced a supervised regularization trick to optimize the SD model effectively.

We investigated the performance of our model on four independent public dialogue datasets with complex human policy. The result showed that our model performed better in predicting next-turn actions than SL and OIL baselines. Meanwhile, we empirically demonstrated that our SD model tackles the covariate shift problem better by comparing action prediction performances in the first and second half of dialogues.

\section{Related Work}
\subsection{Policy Learning}
SL has been utilized in PL for several decades and exhibited great ability \cite{ni-2022-recent,henderson-2008-hybrid}. Recent SL-based studies usually proposed end-to-end models tackling PL and natural language generation tasks simultaneously. For example, \citet{lubis-2020-lava} utilized a variational auto-encoding method to optimize dialogue policy and generate responses; \citet{sun-2022-mars} proposed a contrastive learning method for task-oriented dialogue. However, these studies did not investigate whether their models can replicate human action accurately. RL-based studies typically assume a user has an explicit goal when a dialogue starts. They investigated how to ask the most appropriate problem to know the user's goal as quickly as possible. These studies generally require an agenda-based user simulator \cite{li-2016-user}. For example, \citet{deng-2021-unified} proposed a graph-based model for PL; \citet{ren-2021-learning} improved model performance by introducing an external knowledge graph. Recently, \citet{ouyang-2022-training,openai-2022-chatgpt} adopted the prompt method and RL framework to construct an end-to-end general dialogue system. These models obtained astonishing performance in answering user questions (response to user prompts).

\subsection{Imitation Learning}
Imitation learning (apprenticeship learning) trains agents via expert demonstration datasets, which consist of two categories of methods named behavior cloning and inverse RL, respectively \cite{abbeel-2004-apprenticeship}. Behavior cloning can be regarded as a synonym for SL, while inverse RL trains an agent by optimizing the reward function and policy function alternatively \cite{abbeel-2004-apprenticeship}. As the proposed SD model is a variant of an inverse RL model, we only briefly introduce related inverse RL-based advances here. \citet{ho-2016-generative} proved that the inverse RL training process is equivalent to minimizing the occupancy measure of expert policy and target policy under a regularized function in their GAIL model. However, the original inverse RL model and Ho's GAIL model are infeasible in purely offline settings, as estimating the occupancy measure of a target policy also requires online environment interaction. To solve this problem, \citet{lee-2019-truly} proposed the DSFN that utilized a neural network to estimate the target policy's occupancy measure. Then, \citet{nachum-2019-dualdice} and \citet{kostrikov-2020-imitation} tackle the problem by indirectly estimating the distribution ratio, respectively. \citet{chan-2021-scalable} proposed a variational inference-based model to learn policy, and \citet{jarret-2020-strictly} utilized the energy-based distribution and a surrogate function to learn policies, respectively. All these studies only focus on control problems with a low-dimensional state space. We did not find studies that used OIL in dialogue systems.

\section{Methodlogy}


\subsection{Preliminary}
We represent a dialogue as a model-free MDP without rewards (MDP$\backslash$R) defined by a tuple $(\mathcal{S},\mathcal{A},p_{0},p(s'|s,a),\pi(a|s),\gamma)$, where $\mathcal{S}$ is the state space and $s_t=\{U_0^m,U_0^u,...,U_t^m,U_t^u\}$ is a sample of $\mathcal{S}$ at dialogue turn $t$. $U_t^m$ and $U_t^u$ are utterance token lists of the agent and user at turn $t$, respectively. In other words, we define the dialogue context at turn $t$ as the state. In the following content, we also use $s,a$ and $s',a'$ to represent the current state, current action, next state, and next action, respectively. $a_t \in {[0,1]}^k$ is a $k$-dimensional binary vector that represents an action in action set $\mathcal{A}$. Each action consists of $k$ types of sub-action, and we allow the agent to take multiple sub-actions simultaneously. The utterance order of sub-actions is ignored for simplicity. $p_0$ and $p(s'|s,a)$ represent the initial and state-action transition distribution, respectively. $\pi(a|s)$ represents the policy function. $\gamma$ is the discounted factor.

We train the agent in a minimal setting for ease of deployment. Specifically, the agent cannot interact with simulators, does not require dialogue state annotation, and has no access to external knowledge. Under the MDP$\backslash$R assumption, we represent the expert demonstration dataset as a set of the initial state and state-action-next-state transition tuples $\mathcal{D}={\{s_0^i,s^i,a^i,{s'}^{i}\}}_{i}^{D}$ generated by an unknown expert policy $\pi_e$. $s_0^i$ is the initial state of the state sequence that contains $s^i$.

\subsection{Imitation Learning Objective}
This study learns a policy $\pi\colon \mathcal{S}\rightarrow\mathbb{R}^{|k|}$ to recover $\pi_e$ according to $\mathcal{D}$. To utilize the sequential state transition information, we utilized the occupancy measure $\rho_{\pi} \colon \mathcal{S} \times \mathcal{A} \rightarrow \mathbb{R}$ defined as in Eq. \ref{eq:beta} and Eq. \ref{eq:occupancyMeasure} \cite{nachum-2019-dualdice}:
\begin{equation}
  \begin{aligned}
    \beta_{\pi,t}(s)&\triangleq P(s=s_t|s_0\thicksim p_0, a_k\thicksim\pi(s_k), \\ &s_{k+1}\thicksim p(s_{k+1}|s_k,a_k)) \, 0<k<t,
    \label{eq:beta} 
  \end{aligned}
\end{equation}
\begin{equation}
  \begin{aligned}
    \rho_{\pi}(s,a)&=(1-\gamma)\sum_{t=0}^{\infty}\gamma^{t} [\\ &P(s_t=s,a_t=a|\beta_{\pi,t},\pi)].
    \label{eq:occupancyMeasure}
  \end{aligned}
\end{equation}
The occupancy measure implicitly describes state-transition information in $\beta_{\pi,t}$. The occupancy measure can be interpreted as the distribution of state-action pairs that an agent encounters when navigating the environment with policy $\pi$. \citet{syed-2008-apprenticeship} proved that $\pi$ is the only policy whose occupancy measure is $\rho_{\pi}$, and vice versa. Thus, we can use sequential state transition information and recover the $\pi_e$ via $\pi$ by minimizing the difference between $\rho_{\pi}$ and $\rho_{\pi_e}$. Inspired by the work on  distribution correction estimation framework, we used the Donsker Varadhan representation of Kullback Leibler (KL) divergence and derived the objective function \cite{nachum-2019-dualdice,kostrikov-2020-imitation,donsker-1975-asymptotic}:
\begin{equation}
  \begin{aligned}
    {\rm min} \, &{\rm KL}(\rho_{\pi}||\rho_{\pi_e})=\mathop{\rm min}\mathop{\rm max}\limits_{x\colon \mathcal{S}\times\mathcal{A}\rightarrow \mathbb{R}} [\\ &{\rm log} \mathbb{E}_{(s,a)\thicksim\rho_{\pi_{e}}}e^{x(s,a)}-\mathbb{E}_{(s,a)\thicksim\rho_{\pi}}x(s,a)].
    \label{eq:KLOrigin} 
  \end{aligned}
\end{equation}
$x(s,a)$ is an unknown function. To make Eq. \ref{eq:KLOrigin} can be optimized in practice, we used the variable changing trick proposed in \citet{nachum-2019-dualdice} that defined another state-action value function $\varphi\colon \mathcal{S}\times\mathcal{A}\rightarrow \mathbb{R}$ that satisfies:
\begin{equation}
  \varphi(s,a) \triangleq x(s,a) + \mathcal{B}^{\pi}\varphi(s,a),
  \label{eq:bellman}
\end{equation}
\begin{equation}
  \mathcal{B}^{\pi}\varphi(s,a)=\gamma\mathbb{E}_{s'\thicksim p(s'|s,a),a'\thicksim \pi(s')}\varphi(s',a').
  \label{eq:bellmanOperator}
\end{equation}

According to Eq. \ref{eq:bellman}, the last term of Eq. \ref{eq:KLOrigin} follows \citet{nachum-2019-dualdice}:
\begin{equation}
  \begin{aligned}
    &\mathbb{E}_{(s,a)\thicksim \rho_\pi}x(s,a)/(1-\gamma)\\
     &=\sum_{t=0}^{\infty}\gamma^t \mathbb{E}_{s\thicksim \beta_{\pi,t},a\thicksim\pi}[\varphi(s,a)- \gamma\mathbb{E}_{s'\thicksim p,a'\thicksim \pi} \varphi(s',a')]\\ 
     &=\mathbb{E}_{s_0 \thicksim p_0,a_0 \thicksim \pi}\varphi(s_0, a_0).
    \label{eq:linearTermConvert}
  \end{aligned}
\end{equation}

Then, Eq. \ref{eq:KLOrigin} can be reorganized as:
\begin{equation}
  \begin{aligned}
    {\rm min} \, &{\rm KL}(\rho_{\pi}||\rho_{\pi_e})=\mathop{\rm min}\limits_{\pi} \mathop{\rm max}\limits_{\varphi} [\\ &({\rm log}\mathbb{E}_{(s,a)\thicksim \rho_{\pi_e}}(e^{\varphi(s,a)-\gamma\mathcal{B}^{\pi}\varphi(s,a)}))-\\ &((1-\gamma)\mathbb{E}_{s_0 \thicksim p_0,a\thicksim \pi(s_0)}\varphi(s_0,a_0))].
    \label{eq:KL2nd}
  \end{aligned}
\end{equation}

The quantity of the expected Bellman operator $\mathcal{B}^{\pi}\varphi(s,a)$ can be approximated by using the Fenchel duality or replaced by a sample $\varphi(s',\hat{a}^{'})$ \cite{nachum-2019-dualdice,kostrikov-2020-imitation}. For simplicity, we use the $\varphi(s',\hat{a}^{'})$ to replace $\mathcal{B}^{\pi}\varphi(s,a)$ in this study. Then, we can write the empirical objective function $\hat{J}_{OIL}$ as:
\begin{equation}
  \begin{aligned}
    \hat{J}_{OIL}(\pi,\varphi)&=\frac{\gamma-1}{|\mathcal{M}|}\sum_{i=1}^{|\mathcal{M}|}\varphi(s_0^i,\hat{a}_{0}^i)+\\&{\rm log}\sum_{i=1}^{|\mathcal{M}|}e^{\varphi(s^i,a^i)-\gamma\varphi({s'}^i,\hat{a'}^i)}.
    \label{eq:KLEmipirical}
  \end{aligned}
\end{equation}

$\mathcal{M}$ represents a mini-batch data sampled from $\mathcal{D}$, where $\hat{a}_{0}^{i}$ and $\hat{a'}^{i}$ indicated the sample of $a_{0}^i$ and ${a'}^{i}$ generated by $\pi$, respectively. We utilized a neural network parameterized by $\theta$ and a neural network parameterized by $\nu$ to model $\pi$ and $\varphi$, respectively. Both $\pi_\theta$ and $\varphi_\nu$ are represented as an encoder with a multi-layer perception (MLP). 
\begin{equation}
  \pi_\theta(s) = {\rm MLP}({\rm Encoder}_\pi(s))
  \label{eq:actorNet}
\end{equation}
\begin{equation}
  \varphi_\nu(s,a) = {\rm MLP}({\rm Encoder}_\varphi(s)\oplus a)
  \label{eq:nuNet}
\end{equation}

Neural network language models (e.g., recurrent neural networks or transformers) can be used as the encoder. The Gumbel-softmax reparameterization trick can be used to generate one-hot encoded differentiable samples of $a_{0}^{i}$ and ${a'}^{i}$  \cite{jang-2017-categorical}. However, we canceled sampling and used the stochastic distribution of $a_{0}^{i}$ and ${a'}^{i}$ directly to decrease variance. Experimental results demonstrated the effectiveness of cancel sampling. The design allows it to adopt both a current state and a subsequent state to optimize parameters during the training phase, while it only needs a current state to generate actions during the testing phase.

\subsection{Supervised Regularization}
Although the model introduced in the last subsection can learn policy theoretically, it may fail in practice because it directly accepts natural language as input. Given such high-dimensional and sparse inputs, parameters in the model are extremely hard to optimize \cite{chan-2021-scalable}. To alleviate this problem, we utilize the loss of SL to optimize parameters more effectively (Eq. \ref{eq:slLoss}). The OIL model with SL loss is called SD. The effectiveness of introducing SL was demonstrated in previous studies \cite{ouyang-2022-training}. Finally, we obtain the objective function shown in \ref{eq:finalLoss}.

\begin{algorithm}
  \small
  \caption{SD Training Process}
  \label{alg:train}
  \hspace*{\algorithmicindent} \textbf{Input:} Expert Dataset $D$, learning rate $l$, 
   weight $l_\pi,l_\varphi$ \\
  \hspace*{\algorithmicindent} \textbf{Output:}  $\nu$, $\theta$
  \begin{algorithmic}[1]
      \STATE \textit{Random initialize} : $\nu  $, $\theta$
      \WHILE {$\nu$ and $\theta$ not converge}
          \STATE Sample batch $\{s_0^i,s^i,a^i,{s'}^{i}\}_{i=1}^{\mathcal{M}} \thicksim \mathcal{D}$

          \STATE Compute $\hat{a_{0}^{i}}= \pi_{\theta}(a|s_{0}^{i}) \, \forall i \in\mathcal{M}$
          \STATE Compute $\hat{{a'}^i} = \pi_\theta(a|{s'}^{i}) \, \forall i\in\mathcal{M}$
          \STATE Compute  $\hat{J}(\pi_\theta,\varphi_\nu)$ according to Eq. \ref{eq:finalLoss}
          \STATE Compute gradient w.r.t. $\theta$ and $\nu$.
          \STATE Update $\nu \leftarrow \nu + l_\varphi \times l \times \nabla_\nu \hat{J}(\pi_\theta,\varphi_\nu)$
          \STATE Update $\theta \leftarrow \theta - l_\pi \times l \times \nabla_\theta \hat{J}(\pi_\theta,\varphi_\nu)$
      \ENDWHILE
      \RETURN $\nu$, $\theta$
  \end{algorithmic} 
\end{algorithm}

\begin{equation}
  \begin{aligned}
    \hat{J}_{SL}(\pi_\theta)=&\sum_{i=1}^{|\mathcal{M}|}\sum_{j=1}^{|\mathcal{A}|}y_{ij}{\rm log}(\pi_\theta(s^i)_j) + \\ &(1-y_{ij}){\rm log}(1-\pi_\theta(s^i)_j).
    \label{eq:slLoss}
  \end{aligned}
\end{equation}

\begin{equation}
  {\rm min}\hat{J}(\pi_\theta,\varphi_\nu)=\mathop{\rm min}\limits_{\pi_\theta}(\lambda\hat{J}_{SL} + \mathop{\rm max}\limits_{\varphi_\nu}\hat{J}_{OIL}).
  \label{eq:finalLoss}
\end{equation}

The $\lambda$ is the tradeoff factor. The min-max problem in Eq.\ref{eq:finalLoss} can be optimized by updating $\theta$ and $\nu$ alternatively \cite{ho-2016-model}. The pseudo-code of the optimization is described in Algorithm \ref{alg:train}.

\section{Experiment}
\subsection{Datasets}
We utilized four public datasets to investigate the performance of our model and baselines. (1) Inspired \cite{hayati-2020-inspired}. It is an English movie recommendation dialogue dataset with sociable communication strategies. (2) MultiWoZ 2.3 \cite{han-2021-multiwoz}. It is a corrected version of MultiWoZ 2.0, the largest English multi-domain dialogue public dataset in the world \cite{budzianowski-2018-large}. (3) TGRec \cite{zhou-2020-towards}. It is a Chinese dialogue dataset with topic threads to enforce natural semantic transitions. (4) DuRec \cite{liu-2020-towards}. It is a Chinese dialogue dataset that contains multi-type tasks.

All four datasets were split into train, validation, and test sets when they were released. We selected these datasets because they fully annotated agents' actions at every turn and contained complex social dialogue policy. The semantic meaning of the Inspired and the MultiWoZ 2.3 dataset is significantly more complex than the DuRec and the TGRec dataset because the agents in the DuRec and the TGRec datasets adopt only one sub-action in a single utterance, while agents in the other two datasets may adopt multi sub-actions. Table \ref{table:dataStatistics} described the statistics of the four datasets.

\begin{table}[t]
  \small
  \centering
  \begin{tabular}{lcccc}
      \hline
          & \textbf{Ins.} & \textbf{M.W.} & \textbf{Du.} & \textbf{TG.}\\
      \hline
      \textbf{\# Dialogues}    & 1K &  8K & 10K & 10K  \\
      \textbf{\# Turns}  & 36K   &  114K & 156K & 129K\\
      \textbf{\# Sub-actions}  &  15 & 45 & 24 & 4  \\
      \textbf{Multi sub-actions}      & True & True & False & False \\
      \hline
  \end{tabular}
  \caption{Data Statistics. Ins., M.W., Du., and TG. mean Inspired, MultiWoZ 2.3, DuRec, and TGRec datasets, respectively.}
  \label{table:dataStatistics}
\end{table}

\begin{table*}[tbh]
  \small
  \centering
  \begin{tabular}{lcccccccc}
      \hline
        & \multicolumn{4}{c}{\textbf{Inspired}} & \multicolumn{4}{c}{\textbf{MultiWoZ 2.3}} \\
        & Acc. & AUC & -Log Prob. & APS & Acc. & AUC & -Log Prob. & APS  \\
      \hline
      \textbf{SL} &0.17$\pm$0.02&0.63$\pm$0.02&4.30$\pm$0.11&0.19$\pm$0.01&0.19$\pm$0.01&0.91$\pm$0.01&5.33$\pm$0.15&0.55$\pm$0.01\\
      \textbf{EDM} &\textbf{0.20$\pm$0.01}&0.67$\pm$0.02&5.28$\pm$0.19&0.20$\pm$0.01&0.19$\pm$0.00&0.91$\pm$0.02&5.43$\pm$0.24&0.55$\pm$0.01\\
      \textbf{AVRIL} &0.16$\pm$0.01&0.62$\pm$0.01&4.27$\pm$0.02&0.23$\pm$0.01&0.13$\pm$0.02&0.89$\pm$0.02&5.96$\pm$0.16&0.49$\pm$0.03\\
      \textbf{VDICE} &0.00$\pm$0.00&0.52$\pm$0.01&37.91$\pm$9.21&0.11$\pm$0.01&0.00$\pm$0.00&0.53$\pm$0.01&73.30$\pm$4.22&0.11$\pm$0.00\\
      \textbf{SD} &\textbf{0.20$\pm$0.01}&\textbf{0.71$\pm$0.00}&\textbf{3.95$\pm$0.10}&\textbf{0.27$\pm$0.01}&\textbf{0.23$\pm$0.01}&\textbf{0.94$\pm$0.01}&\textbf{4.82$\pm$0.18}&\textbf{0.60$\pm$0.01}\\
      \hline
      \hline
        & \multicolumn{4}{c}{\textbf{TGRec}} & \multicolumn{4}{c}{\textbf{DuRec}} \\
        & Acc. & AUC & -Log Prob. & APS & Acc. & AUC & -Log Prob. & APS  \\
      \hline
      \textbf{SL} &0.84$\pm$0.00&0.97$\pm$0.00&0.49$\pm$0.01&0.71$\pm$0.00&\textbf{0.91$\pm$0.00}&\textbf{0.99$\pm$0.00}&0.78$\pm$0.03&0.90$\pm$0.01\\
      \textbf{EDM}  &0.85$\pm$0.00&0.98$\pm$0.00&0.58$\pm$0.05&0.72$\pm$0.00&\textbf{0.91$\pm$0.00}&\textbf{0.99$\pm$0.00}&0.78$\pm$0.02&0.90$\pm$0.01\\
      \textbf{AVRIL} &0.86$\pm$0.00&0.97$\pm$0.00&0.49$\pm$0.00&\textbf{0.72$\pm$0.00}&0.91$\pm$0.01&\textbf{0.99$\pm$0.00}&0.84$\pm$0.01&0.89$\pm$0.00\\
      \textbf{VDICE} &0.08$\pm$0.01&0.55$\pm$0.09&14.29$\pm$3.38&0.35$\pm$0.01&0.00$\pm$0.00&0.60$\pm$0.01&37.91$\pm$1.34&0.11$\pm$0.03\\
      \textbf{SD} &\textbf{0.86$\pm$0.00}&\textbf{0.99$\pm$0.00}&\textbf{0.42$\pm$0.01}&\textbf{0.72$\pm$0.00}&\textbf{0.91$\pm$0.00}&\textbf{0.99$\pm$0.00}&\textbf{0.64$\pm$0.03}&\textbf{0.92$\pm$0.00}\\
      \hline
  \end{tabular}
  \caption{Next Action Prediction Performance}
  \label{table:mainPerformance}
\end{table*}

\subsection{Baselines}
We adopted an SL-based model and three recently proposed OIL models as baselines. (1) SL (PLM + MLP). It utilizes a pretrained language model (PLM) to learn state representations and follows an MLP to make predictions. Although simple, the PLM + MLP framework was adopted in many recent models that obtain state-of-the-art (SOTA) performance in the PL task, e.g., Galaxy, UniConv \cite{le-2020-uniconv,he-2022-galaxy}. (2) EDM \cite{jarret-2020-strictly}. It utilizes an energy-based distribution matching model to imitate the demonstrator. (3) AVRIL \cite{chan-2021-scalable}. It learns an approximate posterior distribution in a completely offline manner through a variational approach. (4) VDICE \cite{kostrikov-2020-imitation}. It transforms the objective function of distribution ratio estimation and yields a completely off-policy imitation learning objective. We did not compare RL-based models because they require a reward function while designing a reward function is a complex task and is beyond the scope of this study.

\subsection{Implementation Details}
We used the Deberta-base (for English dialogues) and MacBert-base (for Chinese dialogues) as the encoder because they obtained impressive performance in many natural language processing tasks \cite{he-2021-deberta,cui-2020-revisiting}. The maximum input sequence length was set to 512 tokens after tokenization. If a state contains more than 512 tokens, only the latest 512 tokens are reserved. 

We selected four metrics to evaluate model performance in predicting the next actions. They are accuracy, the area under the operating receiver curve (AUC), average precision score (APS), and mean log probability. The accuracy estimates the ratio that a model can predict the turn-specific same action to a human. The optimal Youden's J statistic determined the cut-off point of accuracy. To estimate the stability of models, we independently ran all experiments five times to calculate and report the standard error.

For optimization, we used the Adam optimizer \cite{kingma-2014-adam}. The learning rate was set to 1e-5, and the maximum iteration number was set to 30,000. We conducted training processes with a warmup proportion of 10\% and let the learning rate decay linearly after the warmup phase. We selected these settings because the SL model obtained the best performance under these settings. The SD model contains about 200 million parameters, and a single optimization process requires about 5-10 hours in an Nvidia Tesla V100. The data and source code of this study was released at GitHub\footnote{Source codes are in supplementary in the review phase.}.

\begin{figure*}[tb]
  \centering
  \includegraphics[width=14.96cm,height=4cm]{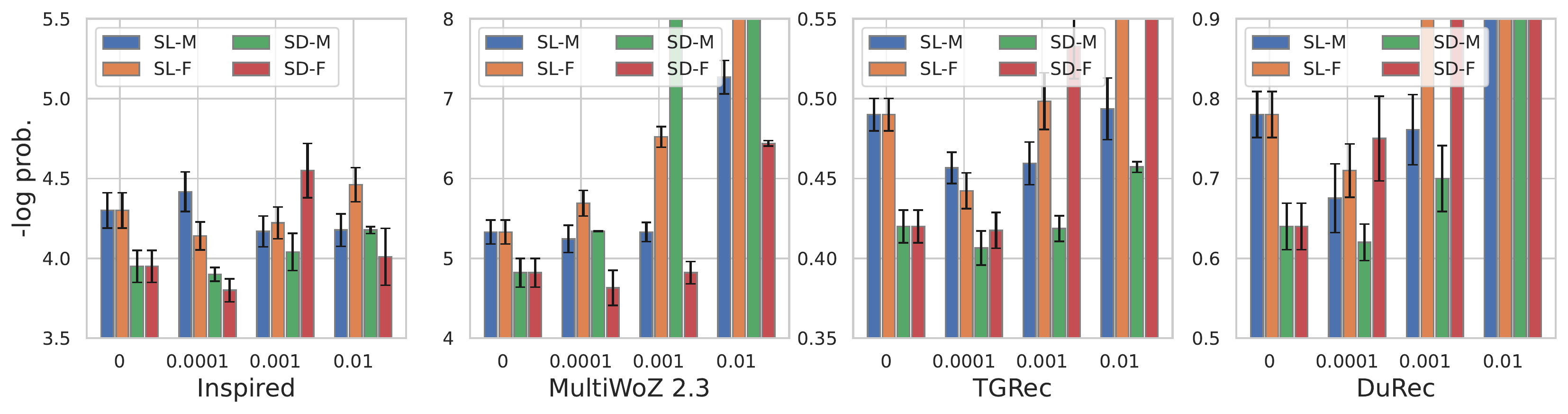}
  \caption{L2-regularized SL Analysis. The SL-M, SL-F, SD-M, and SD-F indicate that we applied l2 regularization to MLP parameters or all parameters for the SL model or the SD model, respectively. 0, 0.0001, 0.001, and 0.01 indicate the weight of the l2 regularization term.}
  \label{fig:overfittingAnalysis}
\end{figure*}

\subsection{Results}
\subsubsection{Action Prediction Performance} 
Table \ref{table:mainPerformance} depicts the action prediction performance. Our SD model performed significantly better than baselines in datasets with more complicated semantic meanings. Specifically, the SD model obtained 0.71 and 0.27 with respect to AUC and APS in the Inspired dataset, while the SL only obtained 0.63 and 0.19, and the three OIL-based baselines only obtained at most 0.67 and 0.23 in the two metrics. The SD model obtained 0.94 and 0.60 concerning AUC and APS in the MultiWoZ 2.3 dataset, while the SL only obtained 0.91 and 0.55, and the three OIL-based baselines only obtained at most 0.91 and 0.55 in the two metrics. The SD model also obtained  significantly better performance in accuracy and log probability (except the accuracy of the Inspired dataset). The SD model obtained modestly better performance in the other two datasets, in which semantic meaning is relatively simple. These results showed that introducing state transition information may improve action prediction performance.
    
Meanwhile, we found that the EDM and the AVIRL only obtained similar performance compared to the SL model, while the VDICE completely failed to converge. Of note, the EDM and the AVRIL used SL information explicitly to train the agent, while the VDICE only utilized SL information implicitly \cite{jarret-2020-strictly,chan-2021-scalable,li-2022-rethinking}. The SD obtained better performance than other OIL baselines, so we only compared SD and SL models in the following content.

\begin{table}[tbh]
  \small
  \centering
  \begin{tabular}{lccc}
  \hline 
  \textbf{Dataset} & \textbf{\# Half} & \textbf{SL} & \textbf{SD}  \\
  \hline
  \multirow{3}*{\textbf{Inspired}} & 1st half  &0.17$\pm$0.03&\textbf{0.19$\pm$0.01} \\ ~ & 2nd half  &0.18$\pm$0.02&\textbf{0.22$\pm$0.01}\\~ & $\Delta$  &-0.01$\pm$0.03&\textbf{-0.02$\pm$0.00}\\
  \hline
  \multirow{3}*{\textbf{MultiWoZ 2.3}} & 1st half &0.21$\pm$0.01&\textbf{0.23$\pm$0.01} \\ ~ & 2nd half  &0.17$\pm$0.01&\textbf{0.22$\pm$0.02}\\~ & $\Delta$  &0.04$\pm$0.01&\textbf{0.01$\pm$0.02}\\
  \hline
  \multirow{3}*{\textbf{TGRec}} & 1st half  &0.87$\pm$0.00&\textbf{0.87$\pm$0.00} \\ ~ & 2nd half  &0.80$\pm$0.00&\textbf{0.85$\pm$0.00}\\~ & $\Delta$  &0.06$\pm$0.01&\textbf{0.02$\pm$0.00}\\
  \hline
  \multirow{3}*{\textbf{DuRec}} & 1st half  &0.92$\pm$0.01&\textbf{0.91$\pm$0.00} \\ ~ & 2nd half  &0.89$\pm$0.01&\textbf{0.90$\pm$0.01}\\~ & $\Delta$  &0.03$\pm$0.01&\textbf{0.00$\pm$0.00}\\
  \hline
  \end{tabular}
  \caption{Half Dialogue Action Prediction Accuracy. $\Delta$ indicates the accuracy differences between the first half and the second half. }
  \label{table:halfDialogueActionPrediction}
\end{table}
\begin{figure*}[tb]
  \centering
  \includegraphics[width=14.96cm,height=4cm]{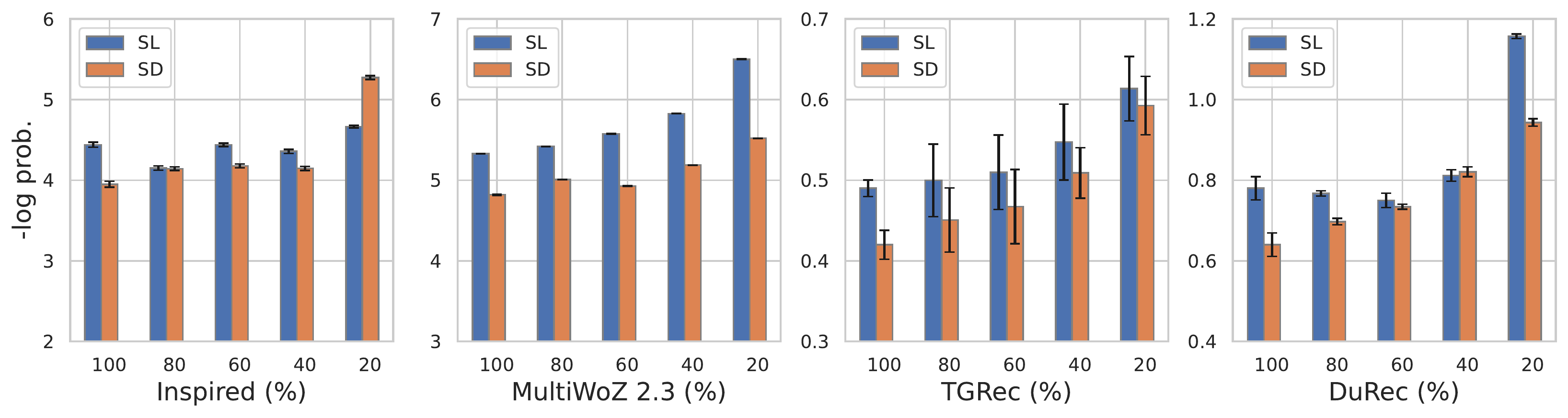}
  \caption{Low Resource Analysis. Numbers in the x-axis represent the used fraction of training data.}
  \label{fig:lowResourceAnalysis}
\end{figure*}

\begin{table*}[tbh]
  \small
  \centering
  \begin{tabular}{lccccc}
  \hline
  \textbf{Model} & \textbf{Inspired} & \textbf{MultiWOZ 2.3} & \textbf{TGRec} & \textbf{DuRec} \\
  \hline
  \textbf{No SL Regularize}  & 15.59$\pm$2.17 &69.03$\pm$4.35 & 4.44$\pm$0.79 & 31.13$\pm$1.73 \\
  \textbf{\quad +SL Regularize} & 5.50$\pm$0.10 &14.03$\pm$0.41 & 0.79$\pm$0.13 & 4.43$\pm$0.27 \\
  \textbf{\quad\quad +Cancel Sampling} & 4.33$\pm$0.11 &10.55$\pm$0.33 & 0.74$\pm$0.02 & 3.13$\pm$0.07 \\
  \textbf{\quad\quad\quad +Full Context} & 4.28$\pm$0.13 &8.13$\pm$0.13 & 0.63$\pm$0.01 & 2.58$\pm$0.02 \\
  \textbf{\quad\quad\quad\quad  +Unlock PLM Parameter} & 4.10$\pm$0.07 &6.45$\pm$0.24 & 0.57$\pm$0.02 & 0.67$\pm$0.05 \\
  \textbf{\quad\quad\quad\quad\quad +Separate Encoder (SD)} & \textbf{3.95$\pm$0.10} &\textbf{4.82$\pm$0.18} & \textbf{0.42$\pm$0.01} & \textbf{0.64$\pm$0.03} \\
  \textbf{\quad\quad\quad\quad\qquad\quad +Larger PLM} & 4.06$\pm$0.07 &5.11$\pm$0.23 & 0.49$\pm$0.04 & 0.68$\pm$0.02 \\
  \hline
  \end{tabular}
  \caption{Ablation Study. Numbers in the table are the -log probability in predicting the next actions.}
  \label{table:ablationStudy}.
\end{table*}

\begin{figure*}[tbh]
  \centering
  \includegraphics[width=14.96cm,height=4cm]{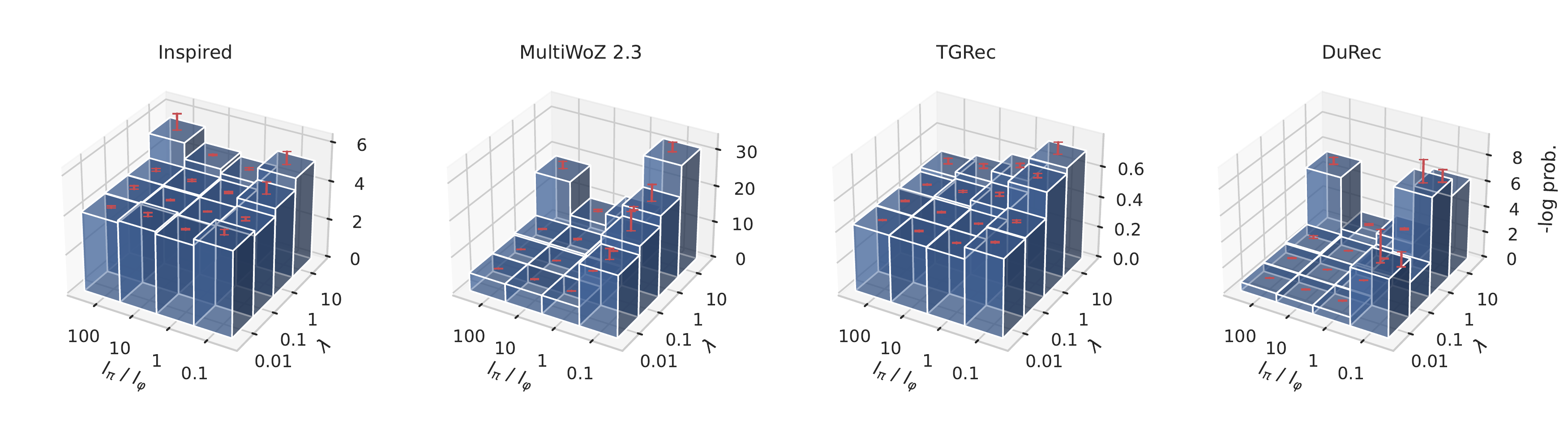}
  \caption{Hyperparameter Analysis}
  \label{fig:hyperparameterAnalysis}
\end{figure*}

\begin{figure*}[tbh]
  \centering
  \includegraphics[width=14.96cm,height=4cm]{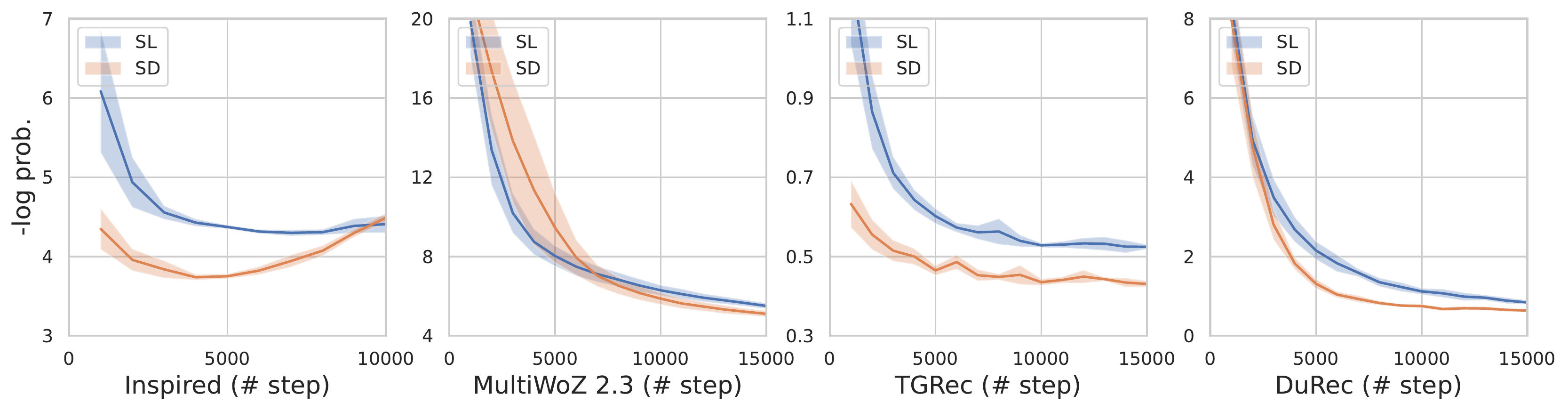}
  \caption{Convergence Analysis. Numbers in the x-axis represent training steps.}
  \label{fig:convergenceAnalysis}
\end{figure*}

\subsubsection{Half Dialogue Action Prediction}
To investigate whether the SD model alleviates the covariate shift problem, we reported the action prediction accuracy of the SD model and the SL model in the first half and the second half of dialogues (Table \ref{table:halfDialogueActionPrediction}). Of note, the statistical difference increases with the development of dialogues. If the SD model indeed alleviates the covariate shift problem, its performance will degenerate slower than the SL model (i.e., $\Delta$ is smaller).

We found that the SD model obtained better performance partly because its performance degenerates slower with the dialogue turns increasing. Performances of both models degenerate in MultiWoZ 2.3, TGRec, and DuRec datasets, while the accuracy difference of our SD model is smaller than the SL model. Specifically, the $\Delta$ of the SL model is 0.04, 0.06, and 0.03 in the three datasets, while the $\Delta$ of the SD model is 0.01, 0.02, and 0.00. Performances of both models improve in the inspired dataset. The accuracy difference of our SD model ($\Delta$ is -0.02) is larger than the SL model ($\Delta$ is -0.01). This finding accords with the assumption that utilizing state transition information can decrease the mistake growth speed \cite{ross-2009-efficient}.

\subsubsection{Regularization Analysis}
Whether OIL models can outperform SL models by utilizing state transition information is a controversial topic \cite{xu-2022-error}. Some recent studies claimed that SL models obtained worse performance because they met overfitting problems during training \cite{li-2022-rethinking}. We investigated models' performances under different l2 regularization settings (Fig. \ref{fig:overfittingAnalysis}). For the space limit, we only describe the performance of -log probability (lower the better) in this and the following experiments.

Experimental results indicate that the SL model met the overfitting problem to some extent. Adding l2 regularization to MLP layers with a small weight (e.g., 0.0001) improves model performance. The l2-regularized SL model obtained the best (average) -log probability as 4.14, 5.24, 0.44, and 0.67 in inspired, MultiWoZ 2.3, TGRec, and DuRec datasets, respectively. The original SL model only obtained 4.30, 5.33, 0.49, and 0.78 in these metrics. However, we surprisingly found that the performance of the SD model also improves by adding l2 regularization appropriately. The l2-regularized SD model obtained significantly better performance than the l2-regularized SL model in Inspired, MultiWoZ 2.3, and TGRec datasets. It also obtained slightly better in the DuRec dataset.

Experimental results of the regularization analysis indicate that our SD model will likely outperform the SL model by utilizing the state transition information rather than avoiding overfitting.

\subsubsection{Ablation Study}
We conducted an ablation study to evaluate the effectiveness of the model design (Table \ref{table:ablationStudy}). The basic version of the SD did not use the supervised regularization term, used the action sampling method, used short context (128 tokens), only optimized parameters of MLPs, and used a shared encoder in $\pi_\theta$ and $\varphi_\nu$. Then, we evaluated model performance by adding the SL loss, canceling the action sampling process and directly using action distribution, using full context (512 tokens), optimizing all parameters in the model, using separate encoders in $\pi_\theta$ and $\varphi_\nu$, and using a larger PLM step by step.

Results show that the SD model cannot be optimized when we do not explicitly utilize supervised regularization. We observed apparent performance improvement when we replaced the action samples with action distributions. Experimental results also indicate that introducing longer context (512 tokens), optimizing parameters of both encoders and MLPs, and utilizing separate encoders to learn text representations benefit performances of the SD model, indicating our design is effective. Meanwhile, we did not observe performance gain by utilizing larger PLMs  such as the Deberta-large.

\subsubsection{Low Resource Analysis}
A PL model needs to obtain acceptable performance with a low resource, as collecting human action annotation is expensive. Therefore, we investigated SD and SL when they were trained by a different proportion of data (Fig. \ref{fig:lowResourceAnalysis}). 
    
Experimental results indicate that our SD model almost always outperforms the SL model when the experimental setting is the same. Specifically, the SD model obtained significantly better performance than the SL model in 18 comparisons and obtained worse performance in only two comparisons. Experimental results also indicate that the SD model can utilize data more effectively. Specifically, the SD model obtained similar performance compared to the SL model by only using about 40\% data in the Inspired, 20\% data in the MultiWoZ 2.3, 40\% data in the TGRec, and 60\% in the DuRec dataset, respectively.

\subsubsection{Hyperparameter Analysis}
Training inverse RL agents is difficult because their performance heavily relies on the choice of hyperparameters. As SD is a variant of inverse RL model, investigating the hyperparameter sensitivity is important to evaluate its usability. 

We investigated the performance of the SD model under different learning rate ratios $l_\pi / l_\varphi$ and SL coefficient $\lambda$. The $l_\pi / l_\varphi \in \{100, 10, 1, 0.1\}$ (larger learning rate was set to 1e-5) and $\lambda\in\{10, 1, 0.1, 0.01\}$. We used the grid search method to choose 16 hyperparameter combinations to train the SD model for each dataset (Fig. \ref{fig:hyperparameterAnalysis}). The performance of the SD model is stable when the value of $l_\pi / l_\varphi$ is between 1 to 100, and the value of $\lambda$ is between 0.01 to 1 in all four datasets. Model performance degenerates significantly when $l_\pi / l_\varphi$ is less than one or  $\lambda$ is larger than 1. These experimental results indicate that our SD model is relatively insensitive to the choice of hyperparameters and has great usability. It converges in a large space.

\subsubsection{Convergence Analysis}
We also investigated the convergence speed of our SD model because inverse RL agents typically require a long time to converge. 

Fig. \ref{fig:convergenceAnalysis} described the convergence speed of SL and SD models, respectively. Experimental results indicated that our SD model generally converged with 4000 to 15000 iterations. The -log probability of SD converged slightly faster than the SL model in the inspired, DuRec, and TGRec datasets and basically the same in the MultiWoZ 2.3 dataset.

\section{Conclusion}
We introduced an SD model consisting of an OIL framework with a supervised regularization component. The SD model can learn complex dialogue policy from human demonstrations, utilize state transition information, and does not require online environment interactions. Results demonstrated that the SD model significantly outperforms SL-based models when dialogue policy is complex and alleviate the covariate shift problem. Our SD model also obtained a similar convergence speed and is relatively insensitive to the choice of hyperparameters. Therefore, we argue that our SD model has great potential to be applied to PL tasks.

\section{Limitations}
We did not presume a user has an explicit goal, while previous PL studies typically presume a user has an explicit goal \cite{lubis-2020-lava,sun-2022-mars}. Therefore, we adopted different metrics to evaluate model performance compared to previous studies. Although we argue that our assumption is appropriate as users do not have a goal in many scenarios, we cannot compare performance between our SD model and many previous PL models.

We empirically demonstrated that the OIL model could obtain better performance. This finding is different from the previous study \cite{xu-2022-error},  which claimed the SL model could learn near-optimal policy in tabular deterministic decision-making tasks. It is important to conduct theoretical analysis to figure out why the OIL model can learn better policy when inputs are raw text and policy is stochastic. However, it is beyond the scope of this study.

Although experimental results demonstrated that our SD model could replicate human policy better than SL and OIL baselines, its performance is still unsatisfying when dialogue policy is extremely complex.

\bibliography{plil}
\bibliographystyle{acl_natbib}
\end{document}